# Faulty Branch Identification in Passive Optical Networks using Machine Learning

**Khouloud Abdelli,**[1,3] * **Carsten Tropschug,**[2] **Helmut Griesser,**[1] **Stephan Pachnicke**[3]

[1]*ADVA Optical Networking SE, Fraunhoferstr. 9a, 82152 Munich/Martinsried, Germany*
[2]*ADVA Optical Networking SE, Märzenquelle 1-3, 98617 Meiningen, Germany*
[3] *Christian-Albrechts-Universität zu Kiel, Kaiserstr. 2, 24143 Kiel, Germany*
*\*Corresponding author: [kabdelli@adva.com](kabdelli@adva.com)*

**Passive optical networks (PONs) have become a promising broadband access network solution thanks to its wide bandwidth, low-cost deployment and maintenance, and scalability. To ensure a reliable transmission, and to meet service level agreements, PON systems have to be monitored constantly in order to quickly identify and localize networks faults and thus reduce maintenance costs, minimize downtime, and enhance quality of service. Typically, a service disruption in a PON system is mainly due to fiber cuts and optical network unit (ONU) transmitter/receiver failures. When the ONUs are located at different distances from the optical line terminal (OLT), the faulty ONU or branch can be identified by analyzing the recorded optical time domain reflectometry (OTDR) traces. OTDR is a technique commonly used for monitoring of fiber optic links. However, faulty branch isolation becomes very challenging when the reflections originating from two or more branches with similar length overlap, which makes it very hard to discriminate the faulty branches given the global backscattered signal. Recently, machine learning (ML) based approaches have shown great potential for managing optical faults in PON systems. Such techniques perform well when trained and tested with data derived from the same PON system. But their performance may severely degrade, if the PON system (adopted for the generation of the training data) has changed, e.g. by adding more branches or varying the length difference between two neighboring branches. etc. A re-training of the ML models has to be conducted for each network change, which can be time consuming. In this paper, to overcome the aforementioned issues, we propose a generic ML approach trained independently of the network architecture for identifying the faulty branch in PON systems given OTDR signals for the cases of branches with close lengths. Such an approach can be applied to an arbitrary PON system without requiring to be re-trained for each change of the network. The proposed approach is validated using experimental data derived from PON system. © 2022 Optica Publishing Group**

## 1. INTRODUCTION

Passive optical networks (PONs) have become an emerging broadband fiber access network solution in the world today thanks to a plethora of benefits offered such as huge bandwidth, cost effectiveness, scalability, flexibility etc. They are renowned as the predominant technology for fiber-to-the-home (FTTH) deployment, providing various communication and multimedia services including high quality triple play service capabilities for data, voice and video, and high-speed internet access, in a cost-effective manner [1]. This is due to the passive nature of the network and the shared fiber infrastructure. Figure 1 illustrates the general configuration of a passive optical network. A PON system is composed of an optical line terminal (OLT) located at the telecom carrier's central office, the optical distribution network (ODN) containing passive optical devices such as splitters, and several optical network units (ONUs) or optical networks terminals (ONTs) installed at the customers' premises providing services. PONs are typically implemented using two prominent technologies: Ethernet (EPON) and Gigabit PON (GPON). PON systems lower the operation-and-maintenance expenses (OPEX) due to the fact that no power requirements or active electronic components, which are more prone to failures in the outside plant, are required [2]. A significant amount of OPEX can be further reduced by implementing and deploying effective proactive or predictive maintenance schemes for fault management in such systems.

PON systems are susceptible to several failures including ODN fiber failures (both the feeder fiber and the distribution fiber could be affected by, e.g. fiber cuts), ONT transmitter/receiver failures, and dirty/cut/bent patch cords or connectors. More than 80% of deployed PON system faults occur within the distribution/drop segments of the network [3]. As reported by the Federal Communication Commission (FCC), more than one-third of service disruptions are caused by fiber-cable problems [4], and such disruptions can lead to tremendous financial loss in business for the service providers or operators [5]. The manual discovery of incidents occurring in PONs requires considerable expert knowledge and probing time until a fault is identified, located, and fixed. This results in a dramatic increase of OPEX and customer dissatisfaction. Therefore, it is beneficial to monitor PON systems in order to automatically, accurately, and autonomously identify and pinpoint network faults, and thereby reduce maintenance costs, minimize downtime, and enhance quality of service.

Monitoring of optical fiber networks has mainly been realized using optical time domain reflectometry (OTDR), a technique based on Rayleigh backscattering. OTDR is widely applied for measuring the characteristics of an installed optical fiber link such as fiber attenuation, length, and connector and splice losses, and for detecting and localizing fiber faults or anomalies within an optical link. An OTDR operates like an optical radar. It injects a series of optical pulses into one of the ends of the fiber under test. Part of these pulses are reflected back towards the source as a result of Fresnel reflections and Rayleigh scattering effect. The amplitude of this reflected signal is recorded as a function of distance. Thus, a recorded OTDR trace (or waveform) illustrating the positions of faults or events along the fiber, is generated, and used for event analysis. OTDR has been widely adopted for monitoring point-to-point optical links. However, applying OTDR to PON systems can be challenging since the backscattered signals from each branch are added together, which makes it very difficult to differentiate between the branches' backward signals [4]. The event analysis is more challenging for the cases of equidistant branch terminations as the reflected signals from the same length branches overlap and add up, which makes it very challenging to identify the branches. As the size of PON systems increases, the number of fiber links to be monitored increments and thereby the complexity of the event analysis rises leading to less reliable monitoring [6]. Furthermore, the large splitting loss induced by optical splitters located at the remote node, results in a significant drop in the backscattered signal (e.g., an 1:32 splitter leads to 30 dB loss in measured power [6]). Hence, a high dynamic range of the ODTR and a highly sensitive fault detection method are needed for accurate event detection and localization. One of the proposed solutions to overcome the aforementioned challenges is to use a tunable OTDR combined with wavelength multiplexers to allow a dedicated monitoring wavelength for each branch [7-10]. However, such a solution is quite expensive due to the high costs of a tunable OTDR instrument. Another solution, which is simple and effective, relies on using optical reference reflectors at the end of each branch to enhance the fault detectability by detecting the presence and height variation of reference reflector peaks [4, 10, 11]. This requires the lengths of the branches to be different, which limits the applicability for real installed networks. Another much simpler solution for identifying the branch in which an optical fault occurs, consists of monitoring the decrease of the optical power received or transmitted at the attached ONU, up to the loss of service in case of a fiber cut. The applicability of such solutions is constrained by the lack of access to ONU performance data in various operator business models.

Recently, machine learning (ML) has emerged as an efficient tool for failure management (e.g., fault identification, failure localization, etc.) in optical communication systems [12-16]. Particularly, ML-based approaches have shown great potential to solve fault management in PONs. In this respect, Usman et al [17] proposed a data-driven approach for fault monitoring in PON systems by combining the use of fiber sensors and ML whereby the fiber Bragg grating sensors having different characteristic grating for each branch are adopted to acquire the monitoring data, and the ML technique is used for fault identification. We presented a gated recurrent unit based autoencoder model that automatically identifies the type of the optical fiber fault in PONs and fully characterizes it without requiring either the intervention of trained personnel or the installation of additional equipment on the network infrastructure [18]. However, such model could not discriminate the faults occurring in similar length branches in PONs. A network-specific approach based on ML techniques, can be adopted for identifying the faulty branch(es) in PON systems while achieving good performance when tested with data derived from that specific network, incorporating normal and faulty branch conditions. However, such an approach fails to perform well if the ML model is applied to new data obtained from the same network after being changed or expanded (e.g., adding more branches, varying the length difference between two neighbor branches, removing some branches, etc.) As a result, every change in the network requires the ML model to be retrained to improve its performance, which is a cumbersome and time-consuming task [19].

To deal with this issue, we propose a generic ML approach, trained independently of the network architecture, for faulty branch identification for the cases of branches with similar (i. e. length difference between two neighboring branches lies between 1 m and 3 m) or different lengths in PON systems by leveraging monitoring data obtained from the reflectors installed at each branch's end. Our method involves first dividing the OTDR traces recorded using a conventional OTDR device from a PON system, where each fiber strand is monitored by an optical reflector, into short fixed-length sequences with a maximum of two reflections, and then applying the ML model to each sequence to produce the faulty branch class. The ML approach is trained to learn different patterns underlying normal or faulty conditions that two close reflections or a single pulse might exhibit in different circumstances, and thus to accurately recognize the different faulty branch scenarios including single faulty branch or several faults. Please note that for this study, the proposed ML model outputs only the identifier(s) of the faulty branch(es) rather than the location of the fault(s) along the faulty branch(es), which can be provided by the measurement method with lumped reflectors. Our approach can be applied to an arbitrary PON system without having to be retrained for each network change. The proposed approach is validated using experimental data obtained from PON system under different circumstances.

This study's main contributions are outlined as follows:
- The proposed model effectively identifies the faulty branch(es) in PON systems even for the cases with close branch lengths (e.g., length difference of 2 m).
- The proposed approach does not require to be re-trained for each change made in the PON system which would save time and effort required for re-training the ML model and collecting new data.
- The proposed approach for monitoring and surveillance in PONs is straightforward and fast; it does not require the installation of additional costly or complex hardware on the network infrastructure (an optical reflector is placed at the end of each branch and a conventional OTDR device is used for recording the OTDR traces which would be split and fed to the ML model for identifying the faulty branches), and the time for recording OTDR measurements (2ms – 2s) is very short, which would speed up the fault diagnosis.

The rest of this paper is structured as follows: Section 2 gives some background information about the ML algorithm involved for building the proposed ML approach and discusses the network-specific method. Section 3 describes the proposed generic model. Conclusions are drawn in Section 4.

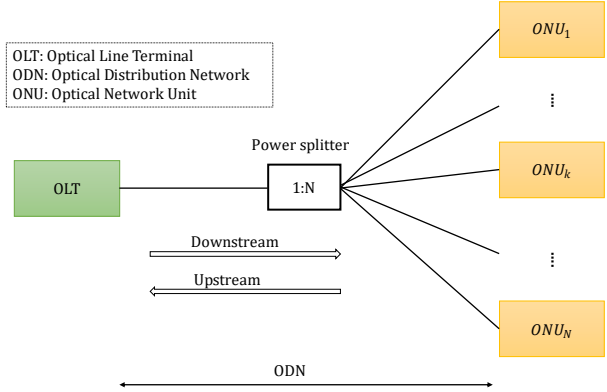

Fig. 1. General architecture of a passive optical network.

## 2. BACKGROUND

This section briefly provides some basic knowledge on the ML technique adopted for the development of the proposed approach.

### 2.1 Long-short Term Memory

LSTM [20] is a particular kind of recurrent neural networks (RNNs) that can handle sequential data, achieving state-of-the-art performance in various sequence classification applications, including speech recognition or natural language processing.

The memory cell, also known as block memory, is the basic computational unit of an LSTM. It is composed of weights and three gates that regulate the information flow to the cell state. The forget gate decides what information to throw away from the cell state. The input gate determines what new information to store in the cell state, and the output gate decides what to output. As shown in Fig. 2, the previous cell state $c_{t-1}$ interacts with the previous cell output $h_{t-1}$ and the present input $x_t$ to determine, which elements of the internal state vector should be updated, kept, or discarded. The LSTM cell is updated by applying the following equations:

$$f_t = \sigma(W_{xf}x_t + W_{hf}h_{t-1} + b_f) \quad (1)$$
$$i_t = \sigma(W_{xi}x_t + W_{hi}h_{t-1} + b_i) \quad (2)$$
$$\tilde{c}_t = \tanh(W_{xc}x_t + W_{hc}h_{t-1} + b_c) \quad (3)$$
$$c_t = f_t \circ c_{t-1} + i_t \circ \tilde{c}_t \quad (4)$$
$$o_t = \sigma(W_{xo}x_t + W_{ho}h_{t-1} + b_o) \quad (5)$$
$$h_t = o_t \circ \tanh(c_t) \quad (6)$$

where $\sigma$ is the logistic sigmoid function, and $f$, $i$, $c$ and $o$ denote the forget gate, input gate, cell activation and output gate vectors, respectively. "∘" represents the Hadamard product operator, all $b$ are learned bias vectors, all $W$ are trainable weight matrices, and $\tilde{c}_t$ is a candidate cell value.

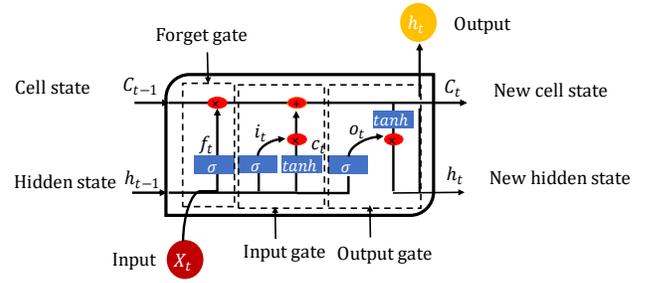

Fig. 2: Structure of a Long Short-Term Memory (LSTM) cell.

### 2.2 Network-specific Approach

The problem of faulty branch identification in PON systems can be solved by adopting a network-specific approach based on ML techniques, in which the ML model is trained using data derived from a PON system, more specifically, long OTDR sequences (i.e., the parts of the OTDR traces after the optical splitter) that include the reflections from all the branches. By learning the patterns of the different faulty scenarios including the circumstances of a single faulty branch or multiple faults occurring at several fiber strands, the ML approach can identify the faulty branch. Figure 3 shows the general architecture of the ML model. The ML approach takes as input OTDR sequences incorporating all the branch reflections, and outputs the faulty branch identifier(s) ($class_{id}$). The length of the input OTDR sequences depends on the PON network topology, particularly the portion of the network following the optical splitter (e.g., number of the ONUs, the length difference between the ONUs, the pulse width and the sampling used for OTDR based PON monitoring, etc.). The sequence length gets longer as the number of ONUs and the distance between branches rises. The number $l$ of classes (i.e., faulty branch scenarios including the cases of a single faulty branch, several faulty branches, all the branches are faulty, etc.) to be investigated depends on the number of branches $m$. $l$ increases with $m$ ( $l \geq 2^m$)). Given that the input of the ML model consists of sequential data (OTDR sequences), recurrent neural networks such as LSTM or gated recurrent unit (GRU) would be a good candidate for solving such problem as they are good at processing sequential data and to capture long term dependencies.

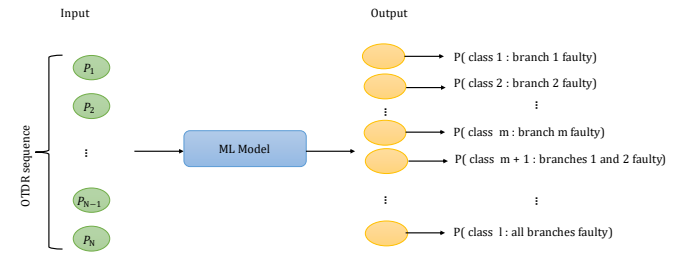

Fig. 3: General architecture for network-specific ML based approach for faulty branch identification in PON systems. The OTDR sequence represents the part of the OTDR trace after the optical splitter including the reflections from all the branches. The output of the ML model is the faulty branch identifier(s).

Such an ML approach performs well given data derived from the same network that was used for the training of the ML model during the learning phase or from a slightly modified network (e.g., longer feeder fiber, lower splitting ratio while maintaining an identical number of branches to the one used to generate the training data, etc.). But if the network is changing (e.g., modifying the length difference between the ONUs, adding, or removing some of the ONUs, etc.), the performance of the ML model may severely degrade as the distinctive reflective pattern that underpins all of the branches in the entire sequence is altered. As a result, for relevant network changes, the ML model must be re-trained. The architecture of the ML approach must be adjusted depending on the change made (e.g., increasing or shortening the length of the input sequence containing the reflections due to all the branches, adjusting the number of classes based on the new number of branches, etc.). However, re-training the ML model for each change in network topology is a time-consuming and inefficient process (considering the time required for collecting new data that includes all the patterns for the investigated classes (the different faulty branch possibilities) particularly, if the number of branches grows, labeling the new data, re-training the model, and deploying the model, among other things). Therefore, it would be beneficial to have a generic ML approach that is trained independently of a specific network topology and that can work with any network architecture without having to be re-trained in case of network changes.

## 3  PROPOSED GENERIC APPROACH

Figure 4 illustrates the workflow of the proposed ML-based approach for faulty branch identification in PON systems. An optical reflector is placed at the end of each fiber strand for OTDR-based PON monitoring. The network topology or architecture of the PON system during deployment or installation is assumed to be known. An OTDR measurement is carried out when the PON is installed or when a network topology change is made. The recorded trace is stored as a reference (pre-measured reference OTDR trace). Once the PON system is deployed, OTDR traces are periodically collected and stored in a centralized database. The recorded OTDR signals are then split into fixed-length sequences. Please note that each OTDR trace starting point for the split comes after the optical splitter position. By considering the pulse width used in the OTDR measurements and the length difference of two closely spaced neighboring branches, the length of the sequence is determined in such a way that the possibility of more than two reflections occurring inside a sequence is extremely low. The generated fixed-length sequences are then fed into the ML model for predicting the type of the investigated event ($C_1$: two reflections where the first reflection is faulty, $C_2$: two reflections where the second reflection is faulty, $C_3$: two reflections, both faulty, $C_5$: one faulty reflection) as well as normal conditions ($C_0$: two normal reflections, $C_4$: one normal reflection, $C_6$: no reflection). The outcomes of the ML model are then compared to the expected outputs in case of normal operation of the PON system (the reference trace). Based on the difference between the predicted outputs and the expected outputs, the faulty branch can be recognized. For example, if the output of the ML model for a sequence is $C_2$, and based on the reference, for that sequence two branches $a$ and $b$ exist, it can be concluded that $b$ is the faulty branch. Please note that the ML approach is trained to learn the patterns of all possible fault scenarios including the occurrence of two simultaneous failures, by taking as input short sequences that cover no more than two reflections. Therefore, the ML model can identify multiple faulty branch situations that may arise from, for instance, a cut in the fiber including several fiber strands between network nodes since the splitting of the OTDR trace recorded under such circumstances, specifically the portion including the reflections from all the branches, would produce short sequences with either maximum two failures or just one fault, depending on the length difference between the branches, which can be accurately identified by the ML model.

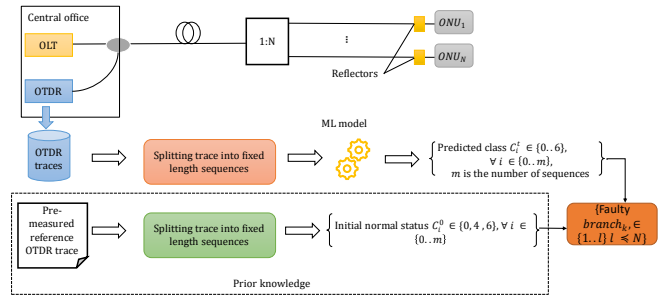

Fig. 4: Flow chart of the proposed approach for identifying the faulty branch in PON systems given OTDR monitoring data. The reference trace represents the OTDR measurement performed when the PON network is deployed or just after making a modification in the network architecture.

### 3.1 Experimental Setup

To validate the proposed ML approach, the experimental setup shown in Fig. 5 is carried out to record OTDR traces incorporating faulty branch and normal operation patterns under different conditions. A real passive optical network is reproduced by adopting a cascade of optical splitters leading to a splitting ratio of 1:128. A reflector is placed at the end of each branch to induce a reflection. Optical attenuators with various attenuation settings (3 dB … 8 dB) are used to adjust the height of the reflections from some branches and thus to model faulty branch conditions. The used attenuators are fixed attenuators with angled physical contact (APC) connectors as they perform better than the variable optical attenuators with physical contact (PC) connectors that could lead to additional reflections. Please note that the placement of the attenuators is carefully chosen to generate the patterns modelling the different investigated classes ($C_i\ \forall i \in \{1 \ldots 6\}$) as illustrated in Fig. 6. The attenuation of the different attenuators is varied from 3 dB to 8 dB. For producing samples incorporating patterns for the class $C_0$, OTDR signals recorded under normal condition (OTDR traces generated using the experimental setup while removing all the attenuators) are adopted. The difference in length of two close branches varies from 2 m to 6 m to generate samples that also include reflected pulses which are very close or even overlap (challenging cases). The OTDR configuration parameters, namely the pulse width, the wavelength and the sampling time are set to 10 ns, 1650 nm and 2 ns, respectively. The pulse width of 10 ns corresponds to a spatial resolution of about 2 m. The laser power is modified from 0 to 16 dBm to influence the SNR. The total measurement time required for recording an OTDR trace after both collecting and averaging several OTDR measurements is varied from 2 ms to 2s. It is to be noted that the total measurement time for generating an OTDR signal to be fed into the ML model for analysis, is very short which will aid in hastening the diagnosis of faults in PON systems. Please note that the ML model works well even when given noisy OTDR signals recorded for short periods of time because it is capable of learning and extracting knowledge from noisy data and the noise could even help to avoid the overfitting of the ML model. This contrasts with the conventional OTDR-based event analysis method, which requires a lot of averaging of OTDR signals to reduce the noise and thus to perform well. Figure 7 shows the normal and faulty branch behaviors of the branches at different attenuation settings.

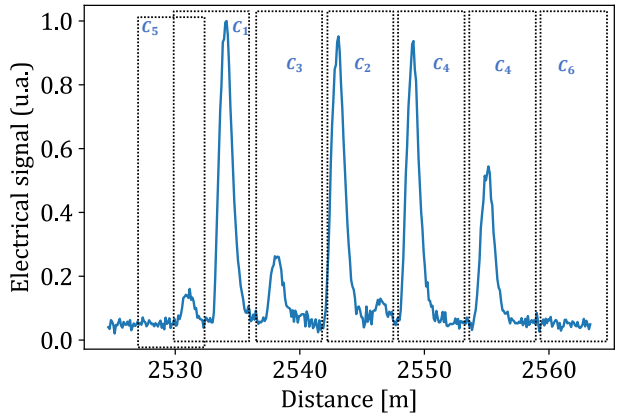

Fig. 6: Example of a generated OTDR signal incorporating the different patterns of the classes to be investigated by the generic ML model. Every dashed rectangle (i.e., a sequence of length 60 extracted from the OTDR trace) represents the pattern of a class $C_i$, $i \in \{0 \ldots 6\}$. The first and the second dashed rectangles model two distinctive patterns by varying the start of split index. The first dashed rectangle incorporates the pattern of $C_5$ whereby only one faulty reflection exists. The second rectangle represents the pattern of $C_1$ where the first reflection is faulty.

### 3.2 Data Preprocessing

The generated OTDR signals, particularly the parts of the traces after the 1:8 splitter, are split in short sequences of length 60. Please take note that we set the length of the sequence to 60 to have a maximum of two reflections within a sequence because, for our studies, a single reflection can be represented by a sequence of length 30 for the selected pulse width. Although the settings of the attenuators have been varied, the range of the considered attenuation values from 3 dB to 8 dB is narrow and the resulting data is not diverse enough to build a robust and reliable ML model and could lead to an overfitting of the ML model. Changing the attenuator values is a time-consuming and manual process. Therefore, the height of the reflection of the branch(es) to be considered as faulty within a sequence is artificially reduced. The reduction ratio of height reflection of the considered faulty branch is randomly varied from $\frac{1}{50}$ to $\frac{1}{2}$ to model the potential variation of the reflection due to a wide range of attenuation settings including the cases of extremely high attenuation values that may cause the reflection to disappear completely (a similar effect is observed in case of a fiber cut occurring after the remote node with high splitting ratio or a long feeder fiber). For the faulty scenarios modeled by two reflections within a sequence, the height of the reflection due to a faulty branch is artificially altered while keeping the other reflection as it is under normal operation condition, or the heights of both reflections resulting from two

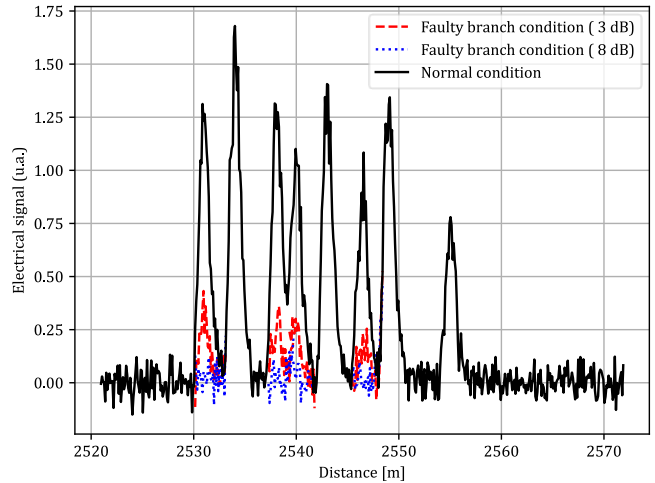

Fig. 7: Normal and faulty branch patterns for different attenuation settings.

faulty branches are reduced simultaneously, arbitrary, and differently, in order to produce diverse data containing various reflection levels that could result from a variety of attenuation settings, including the cases of extremely high attenuation levels that could make either one of the reflections or both reflections totally disappear. Diverse sequences incorporating various patterns for the different faulty scenarios are generated. The normal samples are derived from the experimental setup that was performed without the addition of any attenuator. Given that the generated data is somewhat noisy, additive white Gaussian noise is added to each sequence in such a way that the peak signal-to-noise ratio (PSNR) ranges between 5 and 30 dB (the noise functions as a form of regularization to enhance the robustness of the model). The PSNR is defined as:

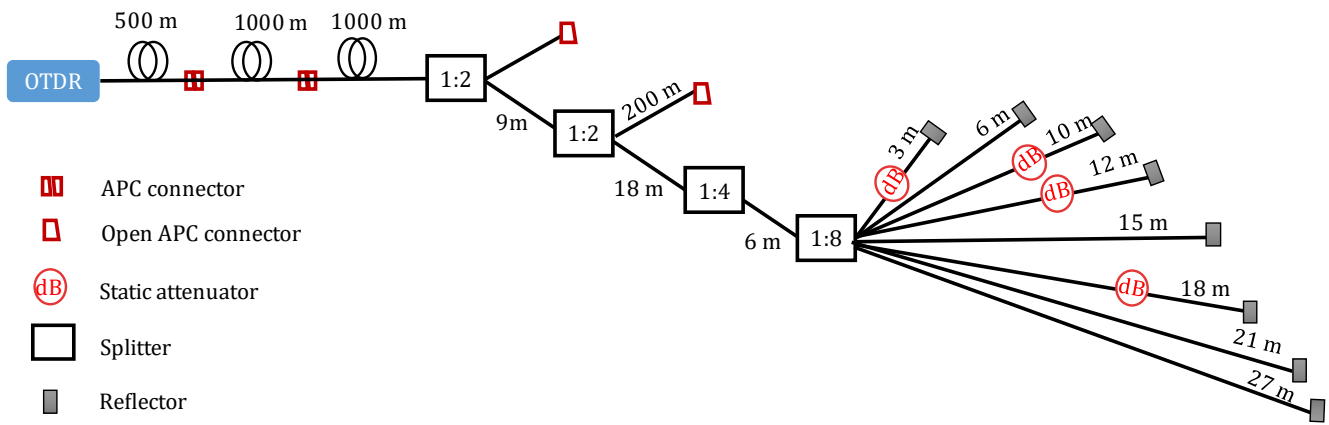

Fig. 5: Experimental setup for generating faulty branch data in a passive optical network.

$$PSNR = \frac{a}{\sigma_{noise}} \quad (7)$$

where $a$ is the height of the reflection and $\sigma_{noise}$ is the standard deviation of the noise.

Figure 8 summarizes the adopted process for generating an example of noisy diverse $C_1$ samples where the first branch (branch 1) is faulty given normal (non-faulty) signals.

In total, a dataset composed of **167,020** samples (~23860 examples for each investigated class) is built, normalized, and divided into a training (60%), a validation (20%) and a test dataset (20%).

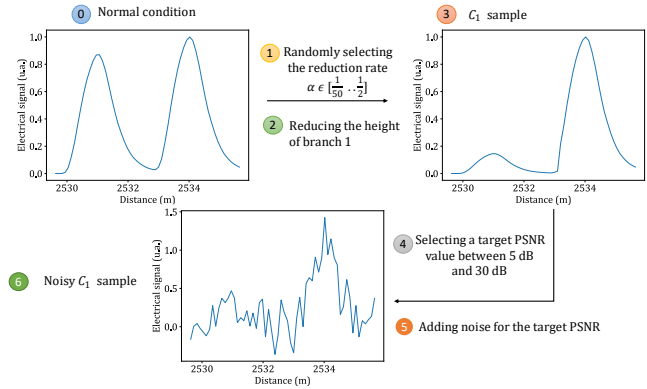

Fig. 8: Process for generating noisy diverse $C_1$ samples (here for branch 1 as an example) given the normal condition signals.

### 3.3 Model Architecture

The architecture of the proposed ML model is illustrated in Fig. 9. The ML approach takes as input a sequence of an OTDR trace of length 60 and outputs the type of class ($C_i \; \forall i \in \{0 \dots 6\}$). The input is fed to an LSTM layer composed of 32 cells, which extract the relevant features $[h_1, h_2 \dots h_{32}]$. The extracted features are then transferred to a fully connected layer containing 16 neurons, followed by an output layer accompanied with a Softmax activation function which can be expressed as follows:

$$\text{Softmax}(z)_i = \frac{\exp(z_i)}{\sum_{j=1}^{k} \exp(z_j)} \quad \text{for } i = 1 \dots k \quad (8)$$

where $z = (z_1 \dots z_k) \in \mathbb{R}^k$ denotes the input vector and $z_i$ represents an element of the input $z$.

The Softmax function outputs the probability distribution over the different classes (the different outcomes of the model).
The model is trained by minimizing the categorical crossentropy (i.e., the cost or loss function) by adopting the Adaptive moment estimation (Adam) optimizer.

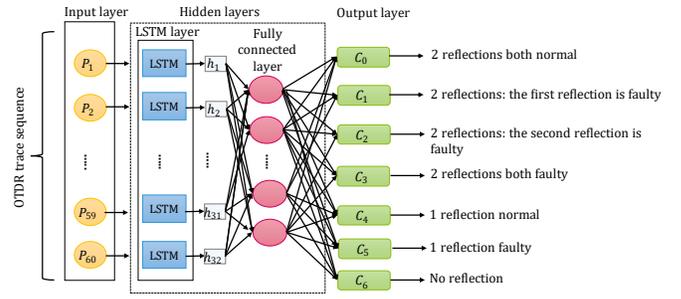

Fig. 9: Proposed generic ML based approach for event recognition.

### 3.4 Performance Evaluation

The performance of the ML approach is first evaluated using the unseen noisy test dataset (randomly derived from the data generated by following the principle explained in subsection 3.2). The confusion matrix depicted in Fig. 10 shows that the ML approach classifies the different classes by achieving a good average diagnostic accuracy of 93%. The ML model might misclassify infrequently the classes $C_1$ and $C_2$ (2 reflections either the first or the second reflection is faulty) as class $C_4$ (one normal reflection) due to the similarity of those classes specifically for the cases where the faulty reflection (one of the two reflections) fades away or is completely gone. The classes $C_5$ and $C_3$ might be misclassified rarely and precisely for the class $C_3$ scenarios where one of the faulty reflections has completely vanished (in such cases, the patterns of $C_3$ look quite similar to $C_5$). The class $C_6$ might be misclassified as $C_5$ for the cases where the height of the reflection is dramatically reduced. The very small misclassification rates yielded by the model can be explained by the fact that under low SNR levels and due to the noise dominating the sequence, the patterns of the investigated classes might look similar due to the noise overwhelming the signal, and thus making it tricky to distinguish between the different classes.

|  | $C_0$ | $C_1$ | $C_2$ | $C_3$ | $C_4$ | $C_5$ | $C_6$ |
|---|---|---|---|---|---|---|---|
| $C_6$ |  |  |  | 0.023 |  | 0.029 | 0.948 |
| $C_5$ |  |  | 0.006 | 0.058 | 0.024 | 0.885 | 0.027 |
| $C_4$ | 0.002 | 0.053 | 0.058 |  | 0.878 | 0.009 |  |
| $C_3$ |  | 0.001 |  | 0.956 |  | 0.032 | 0.011 |
| $C_2$ | 0.005 |  | 0.934 |  | 0.058 | 0.002 |  |
| $C_1$ | 0.005 | 0.925 |  | 0.001 | 0.069 |  |  |
| $C_0$ | 0.977 | 0.009 | 0.013 |  | 0.001 |  |  |

Predicted label / True label

Fig. 10: The confusion matrix achieved by the generic ML approach given the noisy generated test dataset.

The performance of the ML model is assessed then using experimental data without including any artificial noise. The experimental setup used for producing the training data for the generic model is modified to generate the experimental test dataset and thus to evaluate the robustness of the model. The placement of the attenuators is changed: the attenuators are located now at the 6 m, 10 m, 12 m, 18 m, 21 m and 27 m branches, respectively. For the 10 m and 12 m branches, the set attenuation settings are different from the ones used for the training of the ML model. The termination at the end of the 15 m branch is removed to mimic a fiber break fault.

Figure 11 shows the confusion matrix yielded by the ML approach tested with the experimental data generated using the adjusted experimental setup. As the generated data is barely noisy (less noisy compared to the data fed for the training of the ML model), the ML approach classifies with very high accuracy the different classes while yielding very low misclassification rates.

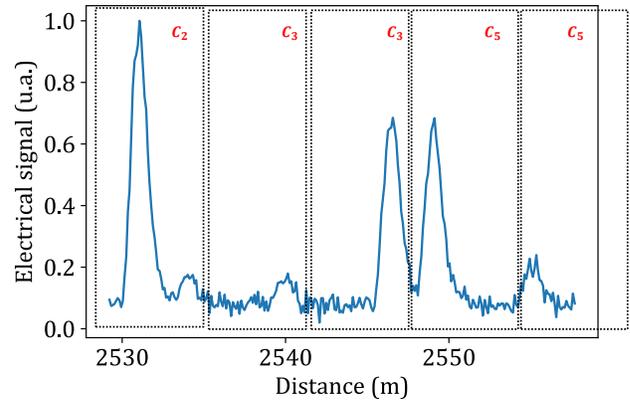

Fig. 12: Example of test OTDR signal split into sequences whereby the outcome of the ML model for each sequence is marked in red.

Fig. 11: The confusion matrix achieved by the generic ML approach given an unseen experimental dataset without adding any artificial noise.

TABLE 1
FAULTY BRANCH IDENTIFICATION GIVEN THE OUTPUTS OF THE ML MODEL AND THE INFORMATION OF THE BRANCH IDENTIFIER

| Output ML Model | Branch Identifier | Faulty Branch |
|---|---|---|
| $C_2$ | 1,2 | 2 |
| $C_3$ | 3,4 | 3,4 |
| $C_3$ | 5,6 | 5,6 |
| $C_5$ | 7 | 7 |
| $C_5$ | 8 | 8 |

### 3.5 Faulty Branch Identification

The outcomes of the ML model along with the knowledge about the PON system topology (the existent branches) are adopted to identify the faulty branches. Figure 12 shows an example of an OTDR signal (a part of an OTDR trace located after the 1:8 splitter) recorded using the adjusted experimental setup described in the previous subsection. The OTDR signal is firstly normalized and split into sequences of length 60. The obtained sequences are fed to the ML model to predict the class of each sequence. Afterwards, based on the outcome of the ML model for each sequence and given the information about the branch identifiers existing in that sequence, the faulty branch can be identified as illustrated in Table 1.

### 3.6 Optimization of the proposed ML model

The best network architecture, attaining the best performance while ensuring a moderate level of complexity, is chosen after evaluation of network architectures of various sizes. The effect of the parameters namely the number of LSTM hidden layers and the number of neurons in the fully connected layer, on the performance of the proposed method in terms of diagnostic accuracy, is investigated. Table 2 shows that a single LSTM hidden layer is the optimum choice for achieving the highest accuracy with the lowest complexity. Table 3 shows that a fully connected layer with 16 neurons is the best choice for striking the ideal balance between accuracy and complexity.

TABLE 2
EFFECT OF THE NUMBER OF LSTM HIDDEN LAYERS ON THE ACCURACY AND TRAINING TIME ACHIEVED BY THE PROPOSED MODEL.

| Number of LSTM layers | Accuracy (%) | Training time (s) |
|---|---|---|
| 1 | 93 | 2522.9 |
| 2 | 92.7 | 4310 |
| 3 | 92.5 | 7844.6 |
| 4 | 92.4 | 14972 |

TABLE 3
IMPACT OF THE NUMBER OF NEURONS IN THE FULLY CONNECTED LAYER ON THE ACCURACY AND TRAINING TIME ACHIEVED BY THE PROPOSED MODEL.

| Number of neurons | Accuracy (%) | Training time (s) |
|---|---|---|
| 8 | 92.4 | 2413 |
| 16 | 93 | 2522.9 |
| 32 | 92.8 | 2990 |
| 64 | 92.9 | 7537.2 |

### 3.7 Comparison of the proposed approach with other ML algorithms

Our proposed ML model is compared with other ML classifiers namely, decision tree (DT), naïve bayes (NB), random forest (RF), gradient boosting (GB), and artificial neural network (ANN). The different ML algorithms are evaluated using an unseen test dataset by adopting the diagnostic accuracy as an evaluation performance metric. The results shown in Fig. 13 demonstrate that our method outperforms the other ML classifiers by achieving the highest accuracy by providing an improvement of more than 2.1% in accuracy as LSTM is better at processing sequential data and to capture long term dependencies.

The comparison results of the inference time (the time required for processing a new test data of 33,404 samples and making prediction) of the proposed model and other ML methods yielded in Fig. 14 (a) show that the proposed method consumes more time in testing (inference) because of the large complexity and parallelism of the LSTM network, while the shallow ML techniques DT, NB, RF and GB require significantly less time for testing.

Figure 14 (b) shows the computational training time required for training each ML model given various input sizes (different number of training samples). The results confirm that the proposed method is slower as it requires more time for the learning phase, and that the other simpler ML models particularly NB, and DT are faster.

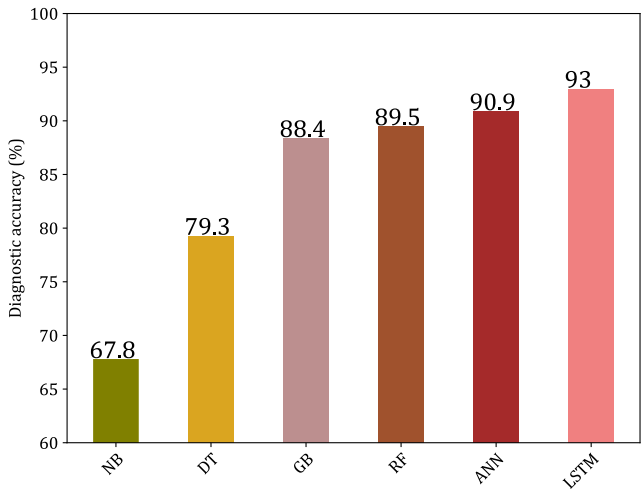

Fig. 13: Comparison of different ML algorithms in terms of diagnostic accuracy.

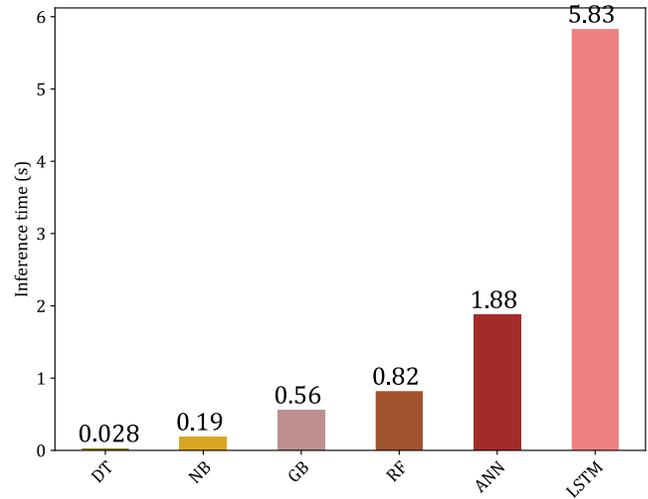

(a)

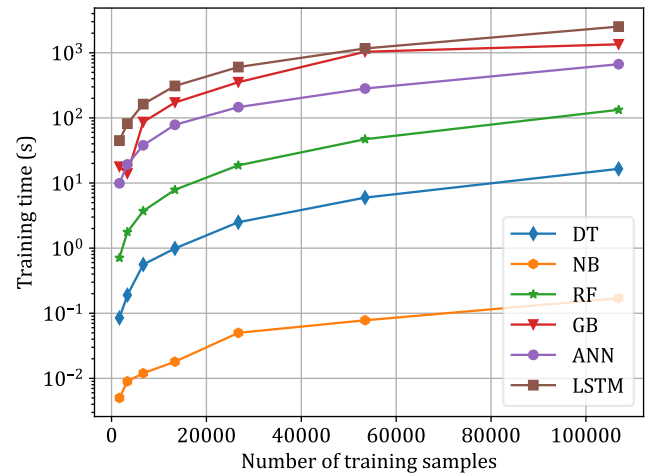

(b)

Fig. 14: Comparison of the computational complexity of different ML models: (a) the inference time of the different models, and (b) the training duration for each ML model as function of the input size (number of training samples).

## 3.8 Applicability for different Network

To evaluate the robustness and the applicability of the ML approach to different PON systems, the experimental setup shown in Fig. 15 is carried out. A PON system is reproduced by using a 1:4 splitter. A variable optical attenuator (VOA) is adopted to vary the attenuation of the reflection and thus to model a faulty branch condition. Some OTDR traces are recorded for different VOA settings for the third (7.3 m branch) and fourth (9.6 m branch) branch. The OTDR parameters (i.e., pulse width, laser power, number of averages of OTDR measurements, sampling time etc.) are the same as for generating the training data. The performance of the ML model is tested given the generated traces, specifically the parts of the OTDR signals after the splitter. Figure 16 shows the outputs of the ML approach for examples of OTDR signals generated for different VOA settings applied for the third and fourth branch. As it can be seen, the ML model could correctly identify the class of each sequence and thus the faulty branch can be correctly recognized, without requiring a re-training of the ML approach with new data derived from the new PON system.

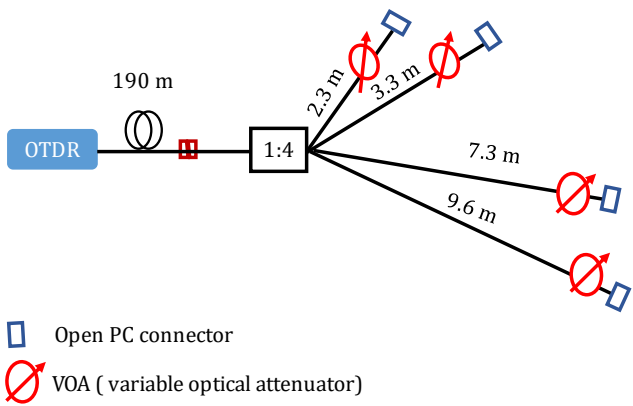

Fig. 15: Experimental setup for a different PON system for testing the robustness of the ML approach.

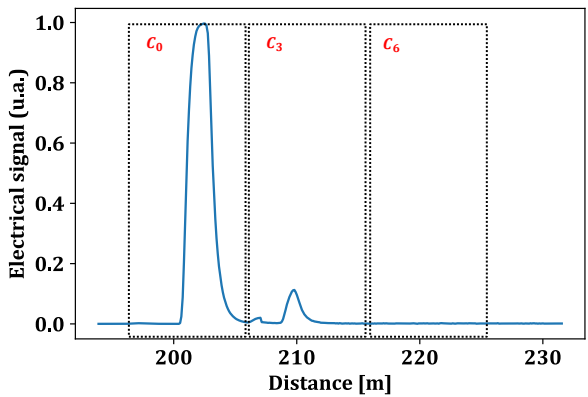

(a)

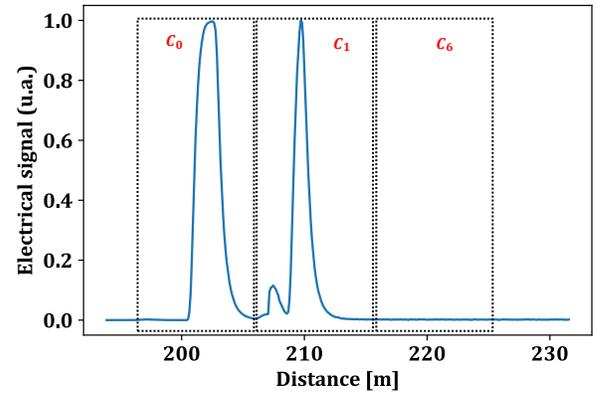

(b)

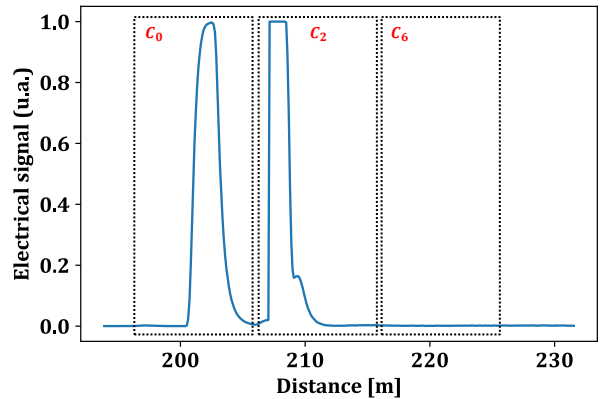

(c)

Fig. 16: Examples of OTDR signals recorded for different VOA settings: (a) 7 dB and 4 dB attenuation set for the third and the fourth branch, respectively, (b) 10 dB attenuation set for the third branch, and (c) 4 dB attenuation set for the fourth branch.

## 4. CONCLUSION

In this paper, an ML-based approach, which is trained independently of a specific network topology, for faulty branch identification in PON systems has been proposed. The proposed method performs well for different PON systems without requiring to be re-trained in case of network changes. The efficiency of the proposed approach has been validated using experimental data derived from PON systems. The presented ML method could help to improve the monitoring of PON networks particularly for the challenging scenarios when reflected pulses overlap. Please note that the proposed approach is trained and tested with sequences of OTDR traces including the whole patterns of the investigated class type. However, in practice, the ML method should be applied to arbitrarily segmented OTDR sequences which might incorporate partial patterns of the events. Therefore, in order to enhance the ML algorithm' adaptability to such probable real-world scenarios, we plan in the future to train it with a dataset that includes both partial and complete patterns for the various investigated classes. In the future, we also intend to extend the architecture of the ML model to output simultaneously the faulty branch identifier(s) and the location(s) of the faults. Furthermore, we plan to enhance further the generalizability capability of the proposed ML approach to address the cases of PON systems with equally spaced branches by re-training the model given data including scenarios of equal-length branches or by applying a transfer learning or continual learning method.

**Acknowledgements.** We would like to thank Mr. Bernd Lebenwein for his support in setting up the experimental setup.
This work has been performed in the framework of the CELTIC-NEXT project AI-NET-PROTECT (Project ID C2019/3-4), and it is partly funded by the German Federal Ministry of Education and Research (FKZ16KIS1279K).